\documentclass[10pt,twocolumn,letterpaper]{article}

\usepackage[]{wacv}   

\usepackage{graphicx}
\usepackage{amsmath}
\usepackage{amssymb}
\usepackage{booktabs}
\usepackage{tabularx}
\usepackage{array}
\usepackage{pifont}

\newcommand{\rom}[1]{\uppercase\expandafter{\romannumeral #1\relax}}
\newcommand{\cmark}{\ding{51}}%
\newcommand{\xmark}{\ding{55}}%

\newcolumntype{F}[1]{%
    >{\raggedleft\arraybackslash\hspace{0pt}}p{#1}}%
\newcolumntype{T}[1]{%
    >{\centering\arraybackslash\hspace{0pt}}p{#1}}%

\usepackage[pagebackref,breaklinks,colorlinks]{hyperref}

\usepackage[capitalize]{cleveref}
\crefname{section}{Sec.}{Secs.}
\Crefname{section}{Section}{Sections}
\Crefname{table}{Table}{Tables}
\crefname{table}{Tab.}{Tabs.}

\def\wacvPaperID{77} 
\def\confName{WACV}
\def\confYear{2025}

\newcommand{\methodName}{UP-FacE}

\usepackage{xcolor}

\begin{document}

\title{\methodName: \underline{U}ser-\underline{p}redictable Fine-grained \underline{Fac}e Shape \underline{E}diting}

\author{Florian Strohm\\
University of Stuttgart\\
Stuttgart, Germany\\
{\tt\small florian.strohm@vis.uni-stuttgart.de}
\and
Mihai Bâce\thanks{Part of this work was conducted while at the University of Stuttgart.}\\
KU Leuven\\
Leuven, Belgium\\
{\tt\small mihai.bace@kuleuven.be}
\and
Andreas Bulling\\
University of Stuttgart\\
Stuttgart, Germany\\
{\tt\small andreas.bulling@vis.uni-stuttgart.de}}

\maketitle

\begin{figure*}[t]
    \centering
    \includegraphics[width=\textwidth]{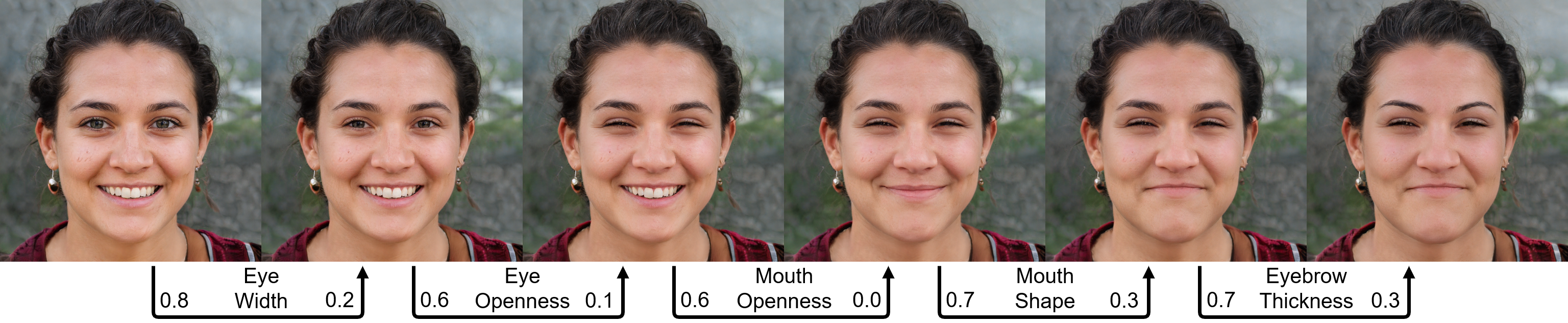}
    \caption{
\methodName~
allows for fine-grained control over various face shape features, such as eye width, eye and mouth openness, or eyebrow thickness.
Qualitative and quantitative results demonstrate that \methodName\ enables precise and fine-grained control over 23 face shape features. 
In stark contrast to existing methods that require trial and error editing of face features, edits with \methodName\ are \textit{predictable} by the human user.
That is, users can control the desired degree of change precisely and deterministically and know upfront the amount of change required to achieve a certain editing result.
In addition, our method enables both isolated progressive (i.e. of the same feature) and sequential (i.e. multiple different features) edits without altering other (unrelated) facial features.  }
    \label{fig:teaser}
\end{figure*}

\begin{abstract}
We present \textit{User-predictable Face Editing (\methodName)} -- a novel method for predictable face shape editing. 
In stark contrast to existing methods for face editing using trial and error, edits with \methodName\ are predictable by the human user.
That is, users can control the desired degree of change precisely and deterministically and know upfront the amount of change required to achieve a certain editing result.
Our method leverages facial landmarks to precisely measure facial feature values, facilitating the training of \methodName\ without manually annotated attribute labels.
At the core of \methodName\ is a transformer-based network that takes as input a latent vector from a pre-trained generative model and a facial feature embedding, and predicts a suitable manipulation vector. 
To enable user-predictable editing, a scaling layer adjusts the manipulation vector to achieve the precise desired degree of change.
To ensure that the desired feature is manipulated towards the target value without altering uncorrelated features, we further introduce a novel \textit{semantic face feature loss}. 
Qualitative and quantitative results demonstrate that \methodName\ enables precise and fine-grained control over 23 face shape features. 
\end{abstract}
\section{Introduction}
Computational face editing is an active area of research with broad applications, such as digital image editing. 
Powerful generative models such as StyleGAN~\cite{karras2019style, karras2020analyzing} can generate high-quality face images and have been shown to learn disentangled latent image spaces that facilitate controlled face editing~\cite{collins2020editing, shen2020interpreting, tewari2020stylerig, wu2021stylespace}.
Because of this, there has been a growing interest in methods that enable users to control the generative process.
Various methods have been proposed for controllable face editing, from unsupervised ones that automatically discover disentangled latent dimensions to control specific parts of the face~\cite{harkonen2020ganspace,shen2021closed,niu2023disentangling} over mask-based methods that use segmentation masks to control the face geometry~\cite{gu2019mask, song2019geometry, lee2020maskgan, ling2021editgan, sun2022ide, sun2022fenerf}, to text-based methods that use natural language~\cite{xia2021tedigan,huang2023collaborative,patashnik2021styleclip,hou2022feat,sun2022anyface,chefer2022image}, 3D-based methods that translate a 3D face model to a real image~\cite{tewari2020stylerig,deng2020disentangled,tewari2020pie,medin2022most,kowalski2020config}, or attribute-based approaches~\cite{choi2018stargan,xiao2018elegant,lu2018attribute,gao2021high,shen2020interfacegan,yao2021latent,abdal2021styleflow,wu2021stylespace} that can control specific appearance attributes. 

While all of these methods can produce impressive results, they also have fundamental limitations that prevent users from easily and deterministically editing the face shape:
Attribute- and mask-based methods are tedious as they require either manual annotations or sufficient skill to manipulate segmentation masks. 
While unsupervised methods overcome these limitations, they are often model-dependent, and dimensions to manipulate the desired attributes might not be discovered or might be entangled with other attributes.
3D-based methods are effective for novel-view synthesis, lighting manipulation, and transferring expressions but require significant 3D modelling efforts for face shape editing.
Most importantly, none of these methods permit users to know the outcome of a particular face edit upfront.
Instead, users must manipulate facial features via trial and error until they are satisfied with the result, resulting in a tedious and error-prone face editing process.

To address these limitations, we present \textit{User-predictable Face Editing} (\methodName).
\methodName~can manipulate 23 face features that describe key characteristics of human face geometry, such as the eye and mouth width and openness, or the chin and eyebrow shape (see \Cref{fig:teaser}).
Similar to action units from the Facial Action Coding System, which describe how face appearance changes for different emotions~\cite{friesen1978facial}, we introduce \textit{semantic face features} that are derived from and describe the relation between facial landmarks and face characteristics.
Unlike existing methods, these semantic features are derived from landmarks, enabling us to quantify the exact facial feature values.
This ability allows us to train \methodName\ without the need for any manual attribute annotations.
To train our model, we propose a novel \textit{semantic face feature loss} that enforces the change in the desired face feature.
Furthermore, we estimated the correlations of the semantic face features on real human faces to dynamically regularise \methodName\, allowing it to modify naturally correlated features jointly.

We report qualitative and quantitative results showing that \methodName\ allows for predictable manipulation of facial features with a high degree of disentanglement.
Most crucially and novel compared to previous methods, our capability to measure and set precise facial feature values enables predictable user editing outcomes: Users can control the desired degree of change precisely and deterministically and know upfront the amount of change required to achieve a certain editing result.
In addition, we developed a user interface that consists of multiple sliders, similar to common digital character editing tools~\cite{schwind2017facemaker, briceno2019makehuman} that provide users with an easy-to-use interface with precise control over the face feature values (see video demonstration: \url{https://youtu.be/xSXAJP1M3ew}), paving the way for more user-friendly digital face editing tools.

\section{Related Work}

\textbf{Unsupervised Face Editing.}
Most unsupervised methods aim at decomposing the latent space~\cite{harkonen2020ganspace,niu2023disentangling,chen2016infogan} or weights~\cite{shen2021closed} of a generative model to identify semantic editing dimensions.
While such methods do not require labelled data, only a limited amount of meaningful semantic directions can be discovered and the level of entanglement is typically higher compared to supervised methods.
DragGAN~\cite{pan2023drag} allows users to edit the shape of a face by moving facial landmarks.
While this allows for a variety of shape edits, this requires significant user effort, especially for edits involving multiple landmarks.
Instead, we specify semantic features using landmarks, which enables \methodName\ to learn how to move these landmarks, thereby facilitating effortless user control.

\textbf{Mask-based Face Editing.}
Mask-based methods condition a generative model on face masks to provide fine-grained control.
One common approach is to allow users to erase parts of the face and sketch the desired outlines of facial features.
Subsequently, a generative model performs image inpainting based on the sketch mask~\cite{portenier2018faceshop, chen2020deepfacedrawing, chen2021deepfaceediting}.
Another approach is to condition face segmentations, where facial features like the eyes and mouth are separated by masks.
This allows users to edit faces by manipulating the face segmentation masks~\cite{gu2019mask, song2019geometry, lee2020maskgan, ling2021editgan, sun2022ide, sun2022fenerf}.
Mask-based methods require significant user effort and skill to manipulate the masks to perform the desired edits.
With our method, a slider interface can be used for face editing, which does not require any special skill.

\textbf{Text-based Face Editing.}
Xia \etal~\cite{xia2021tedigan} and Huang \etal~\cite{huang2023collaborative} combined the ideas from mask-based techniques with text-guided image edits.
Patashnik \etal~\cite{patashnik2021styleclip} combined a pre-trained generator and a powerful CLIP~\cite{radford2021learning} image-language encoder for improved text-guided editing capabilities.
Subsequent work improved upon this to perform more localised edits~\cite{hou2022feat} or to generalise to more diverse text prompts~\cite{sun2022anyface}.
Text-based face editing methods provide a simple language interface to facilitate a large variety of editing possibilities.
However, these methods do not allow for very precise and fine-grained edits.

\textbf{Attribute-based Face Editing.}
Early work on attribute-based face editing focused on binary control of attributes defined
during inference~\cite{choi2018stargan, xiao2018elegant, lu2018attribute, zhang2018generative, guo2019mulgan, yang2021l2m,zhang2018generative, kwak2020cafe, wei2020maggan, he2020pa}, e.g. open or closed mouth.
Other works used attribute classifiers during training for improved editing results~\cite{he2019attgan, sun2022pattgan, gao2021high}.
Generative models conditioned on the attributes have typically limited progressive editing control.
Therefore, Shen \etal~\cite{shen2020interfacegan} trained SVMs to separate face attributes in the generator's latent space with the SVM normal vectors providing global semantic directions along which the corresponding attributes could be manipulated smoothly. 

Another approach is to learn how to modify the latent vectors of a pre-trained generative face model to perform semantically plausible face edits~\cite{yao2021latent, hou2022guidedstyle, khodadadeh2022latent, abdal2021styleflow,wu2021stylespace}.
Yang \etal~\cite{yang2021discovering} proposed a method that did not require binary attributes but sets of images that each shared a particular attribute.
While significant progress has been made in attribute-based face editing, two main disadvantages remain: 
First, the latest methods can only control attributes for which enough manually labelled data exists.
Manual labelling at scale is not only tedious, time-consuming, and costly, but it can also be challenging to provide binary labels for features like \textit{chin shape} for which no clear boundary exists.
Second, while current methods allow for progressive control over attributes, it remains unclear for users how far to move along a semantic dimension to achieve a desired result, e.g. to fully close the mouth.
Our method addresses both limitations as it does not require any manually labelled data and provides semantically meaningful, user-predictable control over the face shape features.
\section{Method}

The core idea of our \textit{User-predictable Face Editing} (\methodName) method is to allow the control of face characteristics through a set of \textit{semantic face features} in a deterministic way.
In the following, we first define our semantic face features and then present how \methodName\ can manipulate the shape of a face based on these features.

\subsection{Landmark-based Face Features}

\begin{table}[t]
\fontsize{9}{10pt}\selectfont
\begin{center}
\begin{tabularx}{1.0\linewidth}{l *2{>{\centering\arraybackslash}X}}
\toprule
Semantic Feature & Landmark Formula  \\
\cmidrule(lr){1-1} \cmidrule(lr){2-2} 
Eye width & $(x_{64} - x_{60}) + (x_{72} - x_{68})$ \\
Eye distance & $(x_{68} - x_{60}) + (x_{72} - x_{64})$ \\
Eye openness & $(y_{66} - y_{62}) + (y_{74} - y_{70})$\\
Pupil position x & $x_{96} + x_{97}$ \\
Pupil position y & $y_{96} + y_{97}$ \\
\midrule
Eyebrow height & $\sum_{i=33}^{50} y_i$\\
Eyebrow width & $(x_{37} - x_{33}) + (x_{46} - x_{42}) \phantom{+}$ \\
Eyebrow thickness & $(y_{41} - y_{34}) + (y_{38} - y_{37}) +$ \\&$(y_{50} - y_{42}) + (y_{47} - y_{45}) \phantom{+}$\\
Eyebrow shape & $(y_{33} - y_{35}) + (y_{37} - y_{35}) +$ \\&$(y_{42} - y_{44}) + (y_{46} - y_{44}) \phantom{+}$\\
\midrule
Nose width & $x_{59} - x_{55}$ \\
Nose length & $y_{57} - y_{51}$ \\
Nose pointiness & $y_{57} - y_{54}$\\
\midrule
Mouth height & $\sum_{i=76}^{88} y_i$\\
Mouth width & $x_{92} - x_{88}$ \\
Mouth openness & $y_{94} - y_{90}$ \\
Mouth shape & $(y_{76} - y_{90}) + (y_{82} - y_{90})$ \\
Upper lip thickness & $y_{90} - y_{79}$ \\
Lower lip thickness & $y_{85} - y_{94}$\\
\midrule
Chin length & $y_{16}$ \\
Chin width & $x_{18} - x_{14}$ \\
Chin shape & $(y_{14} - y_{16}) + (y_{18} - y_{16})$ \\
Jaw width & $x_{23} - x_{9}$ \\
Temple width & $x_{32} - x_{0}$ \\
\bottomrule
\end{tabularx}
\end{center}
\caption{List of the proposed semantic face features based on different facial landmarks.
The characters $x$ and $y$ indicate if a particular landmark's x- or y-coordinate is used, while the subscripts refer to the landmarks shown in \Cref{fig:architecture} (right). 
}
\label{tbl:measure_definitions}
\end{table}

\begin{figure*}[t]
    \centering
    \includegraphics[width=0.9\linewidth]{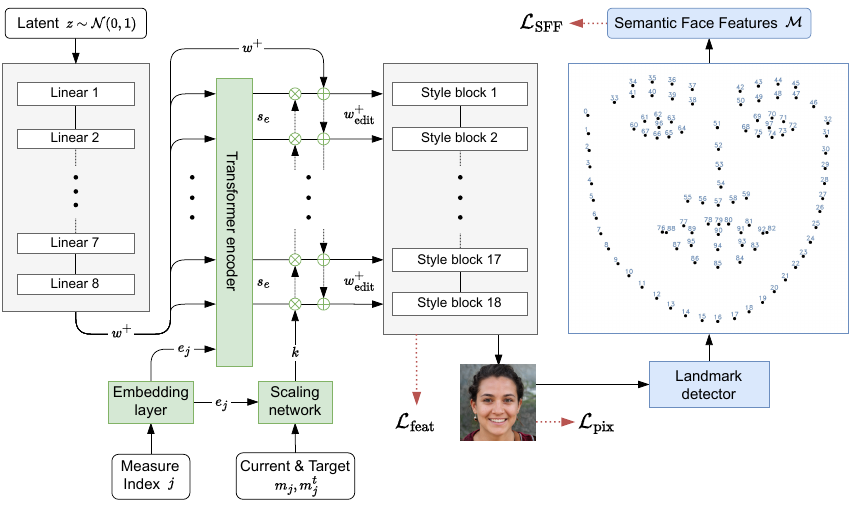}
    \caption{Overview of our method. Shown in grey is the StyleGAN2 architecture. We inject a transformer encoder network between the StyleGAN mapping and synthesis network (green) that can modify the latent vector $w^+$ based on the desired face feature embedding $e_j$ by adding a semantic manipulation vector $s_e$. This manipulation vector is scaled by $k$, a scalar predicted by the scaling network based on the current and target face feature values $m_j$ and $m_j^t$. The landmark detector and calculated face features (blue) are only required during the training of the components highlighted in green. 
    }
    \label{fig:architecture}
\end{figure*}

We define semantic face features $\mathcal{M}$ based on commonly used 98 2D landmarks available in the Wider Facial Landmarks in the Wild (WFLW) dataset~\cite{wu2018look} (see \Cref{fig:architecture} right).
Inspired by common digital face creation tools~\cite{schwind2017facemaker, briceno2019makehuman} that provide an interface with multiple sliders to manipulate semantic face features, we define a total of 23 features $m \in \mathcal{M}$
(see \Cref{tbl:measure_definitions}).
$x$ and $y$ indicate which dimension of a landmark we used for calculating the feature, while the subscripts refer to the corresponding landmark as shown in \Cref{fig:architecture} (right).
The features fall into three different categories: absolute distance, relative distance, and relative anchor distance.
Most features are defined based on the relative distance between two or more landmarks, such as \textit{eye width} or \textit{nose length}.
The advantage of relative distance features is that they are invariant to head translations.
Given a lack of clear reference landmarks, we additionally define five absolute distance features that encode the absolute position of facial landmarks in the face image: horizontal and vertical \textit{pupil position},  \textit{eyebrow height}, \textit{mouth height}, and \textit{chin length}.
The disadvantage of these features is that they are not translation-invariant.
For example, translating the whole face in an image downwards also increases the \textit{chin length}.
However, as we will show later, these features are still effective for controlling the desired facial features.
While it is possible to define relative reference landmarks for these features,
e.g. by relating the landmark used for \textit{chin length} ($y_{16}$) with the position of the mouth, this can lead to undesired side effects because that feature would be affected by both chin length and mouth position. 
For \textit{eyebrow shape}, 
\textit{chin shape}, and \textit{mouth shape}, we define translation-invariant features that relate two landmarks to a third one (anchor) and, as such, allow us to control the angle between them and the shape of the underlying facial feature.
Any differentiable function to combine landmarks can be used, potentially allowing for many more interesting semantic face features. 

\subsection{User-predictable Face Editing}

Our goal is to develop a method that allows the predictable manipulation of the latent vector of an image $I$ in such a way that 
only a desired semantic face feature is changed towards a target value, thus generating a new image $I_\text{edit}$.
\Cref{fig:architecture} shows the overall architecture of the proposed method to achieve this goal.
The grey components are from the state-of-the-art generative network StyleGAN2~\cite{karras2020analyzing} that was pre-trained on the Flickr-Faces-HQ dataset (FFHQ)~\cite{karras2019style} to generate high-quality images of human faces.
A mapping network first maps latent vectors $z \in \mathbb{R}^{512}$ drawn from a standard normal distribution, $z \sim \mathcal{N}(0,1)$, into $\mathcal{W}$ space.
The vector $w \in \mathcal{W}$ is then repeated 18 times, $w^+ \in \mathcal{W}^+ \in \mathbb{R}^{512\times18}$ and input to the generative network that subsequently generates the image.
Prior work has shown that the $\mathcal{W}^+$ space is highly disentangled and can be modulated to perform various semantic image manipulations~\cite{collins2020editing, shen2020interpreting, tewari2020stylerig, wu2021stylespace}.

Inspired by this, we propose to inject a Transformer encoder network~\cite{vaswani2017attention}, highlighted in green in \Cref{fig:architecture}, between StyleGAN2's mapping and synthesis network.
The objective of the Transformer encoder is to predict a semantic manipulation vector $s_e \in \mathbb{R}^{512\times18}$ that is scaled by $k$ and added to $w^+$:
\begin{equation}
\label{eq:w_edit}
w^+_{\text{edit}} = w^+ + k * s_e,
\end{equation}
such that $w^+_\text{edit}$ translates to the same image as $w^+$ except that the value of the selected semantic face feature $m$ is changed to the target value.
We input $w^+$ as a sequence of 18 $w$ vectors with an additional embedding vector $e_j \in \mathbb{R}^{512}$ into the Transformer.
The vector $e_j$ represents an embedding that encodes information about the semantic face feature $m_j \in \mathcal{M}$ to be manipulated.
The embeddings are produced by an additional embedding layer that is trained end-to-end and encodes the 23-dimensional one-hot face feature vectors based on our semantic features defined in \Cref{tbl:measure_definitions}.
The Transformer consists of multiple layers of multi-head self-attention to learn how to manipulate $w^+$ based on $e_j$ to generate $s_e$.
The semantic manipulation vector $s_e$ can be extracted from the Transformer's output by ignoring the last element in the output sequence, which belongs to the input embedding vector.
A scaling network takes the embedding vector and the current and target values for the face feature we want to change, $m_j$ and $m_j^t$, as input.
It predicts a scaling factor $k$ multiplied with $s_e$ as defined in \Cref{eq:w_edit}, allowing us to change the face feature value by a desired amount deterministically.

During the training of the scaling network, embedding layer and Transformer encoder, our method requires a differentiable landmark detector to be able to calculate the face features $M$ (highlighted in blue in \Cref{fig:architecture}).
The weights of StyleGAN2 and the landmark detector model are frozen and only used to calculate the gradient to update the weights of the components highlighted in green.
We define the full loss function $\mathcal{L}$ to train the network as follows:
\begin{equation}
   \mathcal{L} = \lambda_\text{pix} * \mathcal{L}_\text{pix} + \lambda_\text{feat} * \mathcal{L}_\text{feat} + \lambda_\text{SFF} * \mathcal{L}_\text{SFF},
\end{equation}
with $\lambda_\text{pix}$, $\lambda_\text{feat}$ and $\lambda_\text{SFF}$ representing scalars to weight the different loss terms.
$\mathcal{L}_\text{pix}$ is defined as the pixel-based mean squared error (MSE) between the original image $I$ and the modified image $I_\text{edit}$ which aims at preserving the original image as much as possible. 
This loss is zero if the Transformer predicts $s_e =  \overrightarrow{0}$, i.e., the network should perform as few changes as possible on $w^+$ to fulfil the other constraints.
$\mathcal{L}_\text{feat}$ is defined as the feature-based MSE between $I$ and $I_\text{edit}$, where we consider the features of the last style block in the StyleGAN2 generator. 
Similarly to $\mathcal{L}_\text{pix}$, this loss is zero for $s_e = \overrightarrow{0}$ and forces the network to make minimal changes. 
The feature supervision is less rigid than the pixel-based loss and allows the model to be more flexible in changing the face while preserving the original face's appearance.
$\mathcal{L}_\text{SFF}$ is the semantic face feature loss and is responsible for the network to learn how to modify the semantic face features.
It is defined as:
\begin{equation}
\begin{split}
    &\mathcal{L}_\text{SFF}  = \text{ MSE}(m_j^\text{p}, m_j^\text{t}) + \lambda_\text{reg}\ *\\
     &\sum_{i=1, i \neq j}^{23} \text{MSE}(m_i^\text{p}, m_i^\text{t}) * (1 - \lambda_\text{cor}\ *\ |c(m_i, m_j)|).
\end{split}
\label{eq:loss}
\end{equation}
The first term calculates the MSE between the face feature $m_j^p$ of the edited face $I_\text{edit}$ and the target value for the feature $m_j^t$. To calculate $m_j^p$, the generated image is passed through a differentiable landmark detector in the forward-pass as illustrated in \Cref{fig:architecture} to predict the landmarks for $I_\text{edit}$.
While this term encourages the model to change $w^+$ to manipulate the feature $m_j$, we must ensure that all other face features remain unchanged.
Therefore, we add a regularising term with weighting $\lambda_\text{reg}$ that sums the MSE between predicted and target values for all other features where $m_i^t$ is fixed to the original image $I$.
Furthermore, the regularising term is multiplied by the inverse of the absolute Person correlation coefficient $c(m_i,m_j)$ between the feature to be changed $m_j$ and those to be kept fixed $m_i$.
Given that not all of our semantic features are necessarily fully disentangled given human anatomy, we estimate the correlations between the features on the FFHQ dataset~\cite{karras2020analyzing} and introduce scalar $\lambda_\text{cor}$ that controls how much the correlation can relax the regularising term:
If two features are strongly correlated, the regularising term between them is relaxed, and the network is less penalised for changing the correlated features $m_i$ together with the target feature $m_j$.
\section{Experiments}

\subsection{Implementation Details}
Input to our Transformer encoder is the sequence of vectors $w^+ \in \mathbb{R}^{512\times18}$ concatenated with the $512$ dimensional semantic face feature embedding extracted by the embedding layer. The encoder consists of four Transformer encoder layers with four self-attention heads~\cite{vaswani2017attention} each.
The output of the Transformer is input to a linear layer that outputs the semantic manipulation vector $s_e$.
The scaling network takes the current and target face feature values, $m_j$ and $m_j^t$ for a feature $j$, and the feature's embedding $e_j$ as input.
It consists of three linear layers with 32, 32, and one hidden unit(s), respectively, with a ReLU activation function after the first two linear layers and outputs the scaling value $k$.

To train \methodName, we generate data batches by sampling random vectors $z \sim \mathcal{N}(0,1)$ from a standard normal distribution and generate the corresponding vectors $w^+$ and images $I$ using the mapping and synthesis network from a pre-trained StyleGAN2~\cite{karras2020analyzing} model.
For each image, we then sample a face feature index $j$ we want to modify from a uniform distribution $j \sim \mathcal{U}(0,23)$ as well as a desired target feature value $m_j^t$ from a standard normal distribution $m_j^t \sim \mathcal{N}(0,1)$.
Furthermore, we use the pre-trained SPIGA landmark detector~\cite{Prados-Torreblanca_2022_BMVC} to extract landmarks and calculate the current face feature values $m_j$ for each generated image $I$.
Since we sample $m_j^t$ from $\mathcal{N}(0,1)$ we normalise $m_j$ to follow a standard normal distribution as well:
\begin{equation}
    m_j = m_j - \mu_j / \sigma_j,
\end{equation}
where $\mu$ and $\sigma$ are the mean and standard deviation for all face features $j$ calculated on the FFHQ dataset, which our StyleGAN2 model was pre-trained on.
This ensures that our defined semantic face features are normalised to the same value range when calculating the loss.
Using these generated data samples, we can run \methodName\ to predict $w^+_{\text{edit}}$, use the StyleGAN2 synthesis network to generate $I_\text{edit}$ and the SPIGA landmark detector to extract the new landmarks and calculate the predicted feature values $m_j^p$.
We calculate the loss as defined in \Cref{eq:loss} and propagate the gradients back through SPIGA and StyleGAN to finally update the parameters of \methodName.
We optimise \methodName\ for $10^{5}$ steps using the Adam optimiser~\cite{kingma2014adam} with a learning rate of $2^{-5}$ and a batch size of 16.
Through empirical testing, we set the weighting scalars of the loss function to $\lambda_\text{pix}=1$, $\lambda_\text{feat}=3$, $\lambda_\text{SFF}=0.005$, $\lambda_\text{reg}=0.1$ and $\lambda_\text{cor}=1$.

\begin{figure*}[t]
    \centering
    \includegraphics[width=\linewidth]{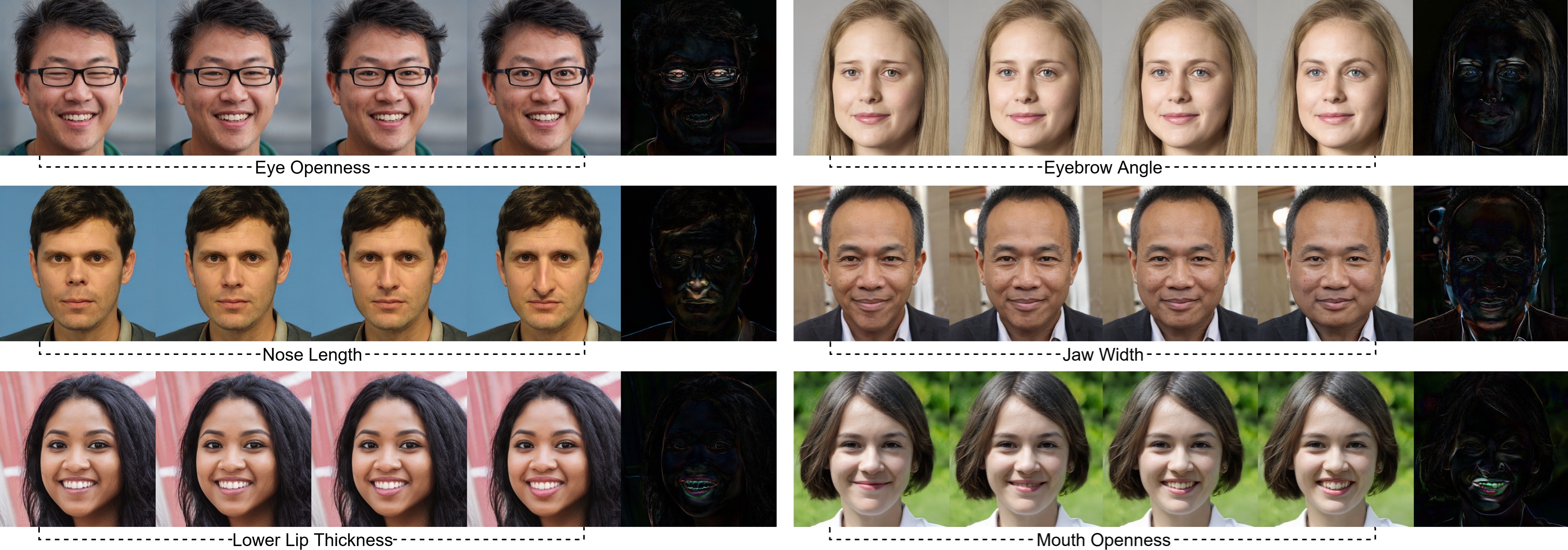}
    \caption{Example progressive edits performed with \methodName. 
    \methodName~allows to easily and deterministically perform high-quality progressive edits along many different semantic dimensions with explicit control over the desired target feature values. For each demonstration of the progressive edits, we also show the difference between the first and last image, highlighting which parts of the image changed. 
    }
    \label{fig:progressive_edits}
\end{figure*}

\begin{figure}[t]
    \centering
    \includegraphics[width=\linewidth]{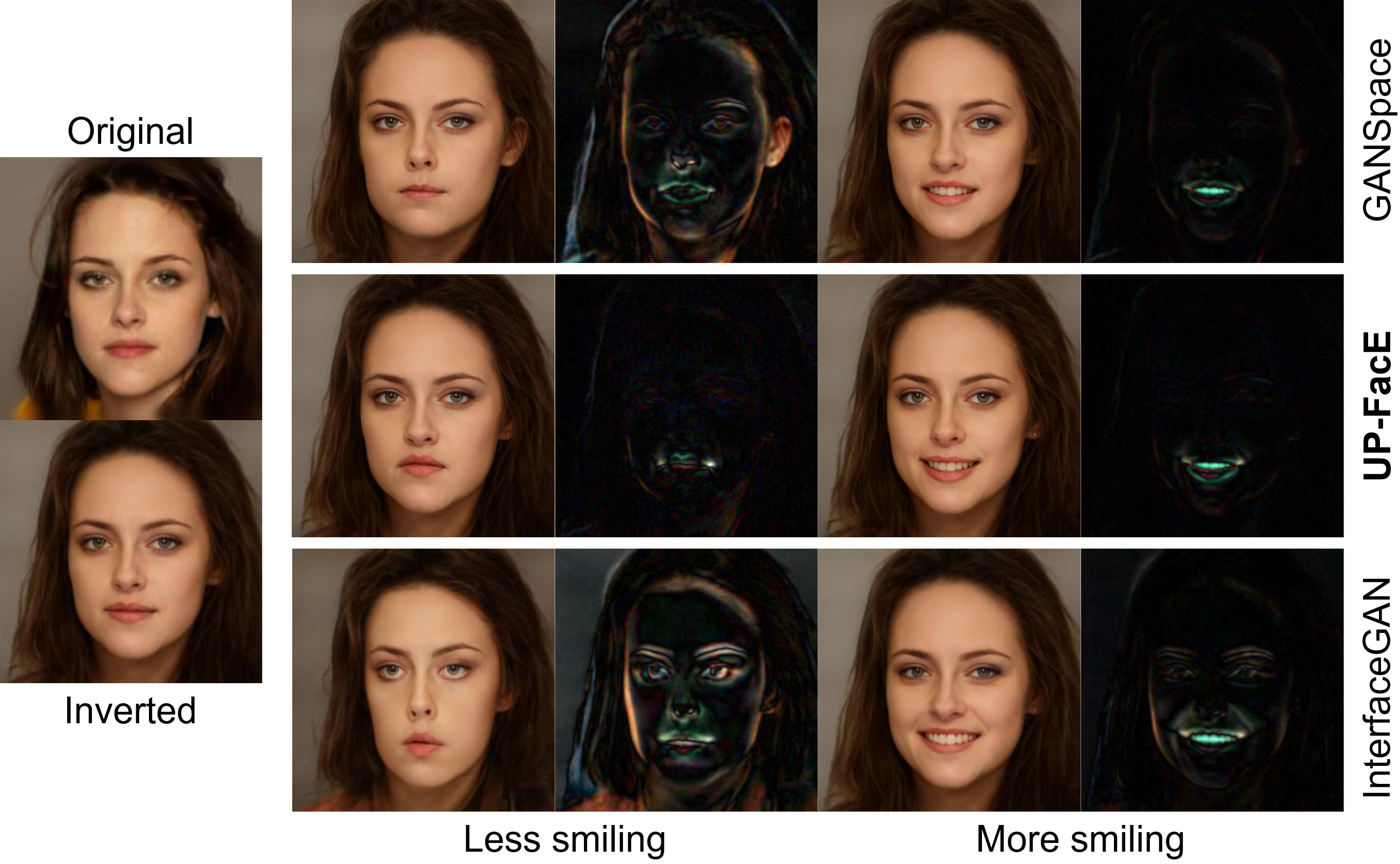}
    \caption{Sample face editing results on a real face image of \methodName~in comparison with the two state-of-the-art methods GANSpace~\cite{harkonen2020ganspace} and InterFaceGAN~\cite{shen2020interfacegan}. The original image was inverted into latent space using the e4e framework~\cite{tov2021designing}, and subsequently, the degree of smiling was edited. Also shown are the difference images between the edited and original images. \methodName~is the only method that allows for fine-grained and deterministic control of the degree of smiling without distortions.}
    \label{fig:real_edit_related}
\end{figure}

\subsection{Qualitative Results}
\Cref{fig:teaser} shows sample sequential edits performed using \methodName.
We first generated a face by sampling a random latent vector and decoded it with StyleGAN2 (left).
Afterwards, we sequentially edited specific features by changing the current value to the desired target value, as indicated by the arrows.
We can see that our method can perform multiple fine-grained edits on the desired features while preserving the visual appearance of the rest of the face, suggesting a high degree of disentanglement.
In stark contrast to prior work, we can calculate the exact value of our face features at any point and set desired values to perform precise and user-predictable editing operations.
This allows users to anticipate the outcome of an edit before its execution, e.g. setting the feature for \textit{mouth openness} to a value of -1.1 always results in a closed mouth when normalised to a standard normal distribution.
Note that it is possible to define reasonable bounds for each face feature and subsequently normalise the values to an arbitrary range as we did for our intuitive slider user interface.
In this case, the smallest value is zero, indicating a closed mouth, and the largest value is one, indicating a fully opened mouth.

Given that our features are on a continuous scale, \methodName\ naturally enables progressive edits as shown in \Cref{fig:progressive_edits}:
\methodName\ can perform the desired changes with fine-grained control over the desired feature values.
For each example, the figure also shows the difference between the first and the last image in the edit sequence.
We can see that \methodName\ operates highly localised, as significant changes are only visible in the relevant regions.
Other changes visible in the difference images are mostly due to high-frequency features, such as the outline of the face.
Minor variations along these lines are visible in the difference images but cannot be noticed in the actual images.

Overall, the examples in \Cref{fig:teaser} and \Cref{fig:progressive_edits} show that our method can perform localised and precise face editing for many facial features.
While the semantic face features defined in \Cref{tbl:measure_definitions} only relate a small number of landmarks to each other, we can model complex face dynamics accurately.
For example, while the definition for \textit{mouth openness} only relates the y-coordinate of two landmarks from the centre of the mouth, the progressive editing results for this feature in \Cref{fig:progressive_edits} show how the whole mouth opens naturally.
We hypothesise that the main reason for the effectiveness of our semantic face features is within the implicit biases and correlations learned by the StyleGAN2 model.
To move the two landmarks used for \textit{mouth openness} apart, \methodName\ learns to open the whole mouth as this is the only possibility given the learned human face distribution of StyleGAN2.

\Cref{fig:real_edit_related} shows editing results on a real face that we inverted into StyleGAN2's latent space using the e4e encoder~\cite{tov2021designing}.
We edited the facial expression twice: once to show less and once to show more smiling.
As the results for \methodName\ in the middle row show, it is also possible to manipulate real faces with high precision, which is generally harder as the inverted latent code of the face might not lay within the well-defined regions of StyleGAN2.
Since smiling is one of the attributes that previous methods can also control, we compare our results with GANSpace~\cite{harkonen2020ganspace} and InterFaceGAN~\cite{shen2020interfacegan}.
We can see that both methods fail to accurately reduce the degree of smiling and start to distort other parts of the face, which is most noticeable in the difference images.
When increasing the degree of smiling, all three methods can accurately edit the face, while \methodName\ still operates more locally.
Additionally, \methodName\ provides more fine-grained control over the mouth as it can edit the \textit{mouth shape} and \textit{mouth openness} separately, allowing us to add a smile without necessarily opening the mouth.
Important to note is that \methodName\ did not require any manually labelled training data, while methods like InterFaceGAN rely on large face datasets with smiling attribute annotations.

\begin{table}[t]
\label{tbl:quantitative_results}
\centering
\begin{tabularx}{1.0\linewidth}{l *4{>{\centering\arraybackslash}X}}
\toprule
Model & Determ.  Editing &  No. of Labels & Accuracy & FID  \\
\cmidrule(lr){1-1} \cmidrule(lr){2-2} \cmidrule(lr){3-3} \cmidrule(lr){4-4} \cmidrule(lr){5-5}
GANSpace & \xmark & 0 & 76.1\% & 36.89 \\
InterFaceGAN & \xmark & 30.000 & \textbf{83.2\%} & \underline{32.58}\\
\methodName & \cmark & 0 & \underline{81.8\%} & \textbf{27.15} \\
\bottomrule
\end{tabularx}
\caption{Quantitative results with two baselines on the smile edit benchmark.}
\end{table}

\subsection{Quantitative Results}
\label{sec_quanti}

We quantitatively evaluate \methodName\ on the smile edit benchmark~\cite{ling2021editgan, lee2020maskgan} and compare its performance against two strong baselines, GANSpace~\cite{harkonen2020ganspace} and InterFaceGAN~\cite{shen2020interfacegan}, using the same StyleGAN2 model for a fair comparison.
The task of this benchmark is to convert faces with neutral expressions into smiling faces.
The performance is measured with three metrics: \textit{attribute accuracy}, which measures whether a face is smiling after editing using an attribute classifier pre-trained on the CelebA~\cite{liu2018large} dataset, \textit{Fréchet Inception Distance (FID)}~\cite{heusel2017gans}, calculated between 4000 edited images and the FFHQ dataset to evaluate image quality~\cite{karras2020analyzing}, and \textit{Identity Score (ID Score)}, which measures if the faces' identity is preserved after editing by calculating the cosine similarity between embeddings extracted from a pre-trained ArcFace model~\cite{deng2019arcface}.
To make a neutral face smile with \methodName\, we adjust \textit{mouth shape} and \textit{mouth openness} as shown in \Cref{fig:real_edit_related}.
The results of the smile edit benchmark are reported in \Cref{tbl:quantitative_results}.
We find that \methodName\ significantly outperforms GANSpace and InterFaceGAN regarding FID and ID scores.
While InterFaceGAN achieves a slightly higher attribute accuracy than \methodName, it is important to note that it was trained on the same 30.000 attribute labels as the classifier used to calculate the accuracy metric.
In contrast, \methodName\ did not require any attribute labels and did not explicitly learn to manipulate the smiling attribute while still being competitive with InterFaceGAN and outperforming GANSpace.
Furthermore, unlike previous methods, \methodName\ allows for precise and predictable control over face shape features.
This could potentially be one reason for the significant improvement in FID and ID scores, as \methodName\ can automatically adjust the editing intensity based on the measured values of the face features.
This allows \methodName\, for example, to only slightly modify faces that are almost smiling, whereas previous methods would move a fixed amount along a semantic editing direction, which can lead to unnecessary over-editing and distortion.

\begin{table}[t]
\fontsize{9}{10pt}\selectfont
\begin{center}
\begin{tabularx}{1.0\linewidth}{l *4{>{\arraybackslash}X}}
\toprule
Model & Pixel \hspace{1em} Error $\downarrow$ & \phantom{ }\newline LIPIPS $\downarrow$ & Edit \hspace{1em} Error $\downarrow$ & Entangle- ment $\downarrow$ \\
\cmidrule(lr){1-1} \cmidrule(lr){2-2} \cmidrule(lr){3-3} \cmidrule(lr){4-4} \cmidrule(lr){5-5}
$\lambda_\text{pix}=0$ & 0.192 & 0.258 & 0.664 & 0.624 \\
$\lambda_\text{feat}=0$ & 0.041 & 0.045 & 0.551 & 0.238 \\
$\lambda_\text{reg}=0$ & 0.043 & 0.047 & 0.549 & 0.267 \\
$\lambda_\text{cor}=0$ & \textbf{0.040} & \textbf{0.041} & 0.575 & \textbf{0.225} \\
\cmidrule(l){2-5}
\methodName & \underline{0.041} & \underline{0.045} & \underline{0.529} & \underline{0.234} \\
\methodName\ 3x & 0.065 & 0.087 & \textbf{0.318} & 0.340 \\
\bottomrule
\end{tabularx}
\end{center}
\caption{Ablation experiments to evaluate the impact of the choice of loss term on different error metrics: pixel $\lambda_\text{pix}$, feature $\lambda_\text{feat}$, regularisation $\lambda_\text{reg}$, and feature correlation relaxation $\lambda_\text{cor}$.
}
\label{tbl:ablation_experiments}
\end{table}

\subsection{Ablation Experiments}

We perform ablation experiments for different versions of \methodName\ and report four different metrics: \textit{pixel error}, \textit{LIPIPS}, \textit{edit error} and \textit{entanglement}.
The \textit{pixel error} is defined as the mean absolute error (MAE) between the original and the modified image.
While we want this to be as small as possible, it should never be zero as this would indicate no change in the face.
\textit{LIPIPS}~\cite{zhang2018perceptual} is a perceptual similarity metric based on neural network features where a smaller value indicates higher similarity.
The \textit{edit error} is defined as the MAE between the predicted feature value $m_j^\text{p}$ and the target $m_j^\text{t}$ for the features $m_j$ that we actually want to change.
Similarly, \textit{entanglement} measures the MAE between $m_j^\text{p}$ and $m_j^\text{t}$ for all features that were not explicitly changed and should stay close to the original value.
We report the results in \Cref{tbl:ablation_experiments}, which we calculated by generating 10,000 random images and then performed edits on five random semantic features for each face.

We can see that the model is unstable during training without the pixel loss ($\lambda_\text{pix}=0$), which is reflected in the lower performance. 
Training without the feature loss ($\lambda_\text{feat}=0$) slightly increases the edit error, indicating that the feature supervision adds some flexibility compared to pure pixel supervision, improving editing precision.
Removing the regularisation term with $\lambda_\text{reg}=0$ results in a higher entanglement, which is expected since we do not penalise the network when changing features other than implicitly through the pixel and feature loss.
Training without the feature correlation relaxation in the regularising term ($\lambda_\text{cor}=0$) significantly increases the edit error as it constrains the model more.
At the same time, we can also see that this model achieves a better entanglement score compared to \methodName, which is expected since this term allows for the entanglement of correlated features in \methodName.
Finally, we compare the results with \methodName\ 3x, which iteratively applies \methodName\ three times by feeding $w^+_\text{edit}$ back as the input to the transformer.
\methodName\ may sometimes over- or undershoot the desired target value for a semantic face feature. 
However, by iteratively applying the method multiple times, the edit error can be significantly reduced, resulting in more precise and user-predictable editing.

\begin{figure}[t]
    \centering
    \includegraphics[width=\linewidth]{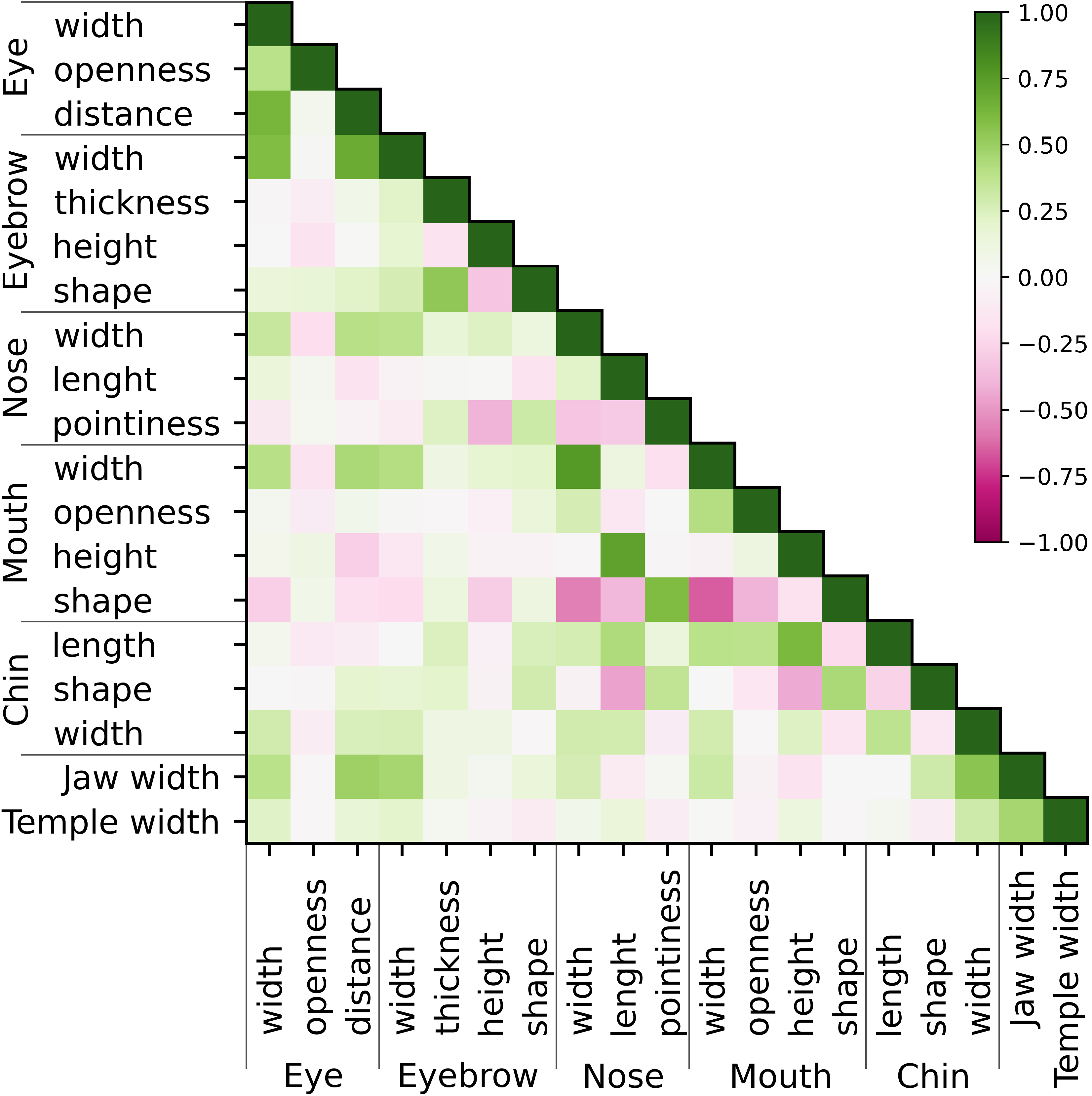}
    \caption{Correlation matrix showing the Pearson correlation between two semantic face features in each cell.}
    \label{fig:correlation_matrix}
\end{figure}

\subsection{Correlation of the Face Features}
The ablation results in \Cref{tbl:ablation_experiments} show that the relaxation of the regularising term based on feature correlations as defined in \Cref{eq:loss} is important to increase edit precision by purposefully entangling correlated features.
\Cref{fig:correlation_matrix} shows the Pearson correlation coefficient matrix for a subset of our features.
We estimate these correlations by calculating all of our semantic face features for each face in the FFHQ dataset.
We can see from the matrix that most features have almost no correlation, while only a small set of pairs are strongly correlated.
For example, \textit{eye width} and \textit{eyebrow width} are strongly correlated, which makes sense intuitively as it is uncommon for humans to have small eyes with very wide eyebrows (and vice versa).
We can see a similarly strong correlation between \textit{mouth width} and \textit{nose width}.
Therefore, by reducing the loss penalty for the network when editing both features together, we give it more freedom to perform the change while ensuring that the edited faces remain aesthetically plausible.
We can also see that \textit{jaw width} is correlated with every other feature related to width.
As an increase in \textit{jaw width} generally relates to a wider face, the other facial features can be scaled accordingly without receiving a high penalty.
Similarly, a wide face with small facial features is uncommon.
Another example of a strong correlation is between \textit{nose length} and \textit{mouth height}, which can be explained by the fact that the distance between the nose and the upper lip, called Philtrum, is typically around 2\,cm with only slight variations~\cite{zankl2002growth}.
Therefore, those two features are strongly correlated and change together, allowing the Philtrum to stay within the natural range.

While most feature correlations can intuitively be explained by natural correlations in human faces, some correlations are likely due to a limitation of our absolute distance semantic face features.
One such example is the correlation between \textit{chin length} and \textit{mouth height}.
Both of these features are absolute distance measures, so they are both sensitive to translations and rotations in y-direction.
For example, if a head is rotated backwards, both of these features will decrease because the chin and mouth move up in the image space, resulting in a false positive correlation.
This suggests that results can be improved even further when using more invariant features in future work.

\section{Conclusion}
We proposed \methodName\ -- a novel method that enables user-predictable and precise face shape editing with little effort.
The key idea behind \methodName\ is the use of 
\textit{23 semantic face features}, determined from different facial landmarks groups.
Unlike existing methods, landmarks have the advantage that they allow the calculation of the values of these face features dynamically, which allows training \methodName\ without requiring any manual labelling.
We also introduced a novel \textit{semantic face feature loss} that encourages the model to manipulate only the desired face features while keeping unrelated features unchanged.
Given that these features are continuous, we trained a scaling network that could learn how to scale the manipulation vector to achieve the desired change in facial appearance. 
This represents a significant advance given that with prior methods, users had to move towards the desired appearance via trial and error.
In contrast, \methodName\ allows for user-predictable face editing, which paves the way for more user-friendly digital face editing tools.

{\small
\bibliographystyle{ieee_fullname}
\bibliography{egbib}
}

\end{document}